\documentclass[conference]{IEEEtran}
\IEEEoverridecommandlockouts
\usepackage{cite}
\usepackage{amsmath,amssymb,amsfonts}
\usepackage{algorithmic}
\usepackage{graphicx}
\usepackage{textcomp}
\usepackage{xcolor}
\usepackage{multirow}
\usepackage{array}
\usepackage{multicol}
\usepackage{etoolbox}
\usepackage{bm}
\usepackage{etoolbox}
\usepackage{caption}
\usepackage{url}
\usepackage{subcaption}

\usepackage{amssymb}
\usepackage{pifont}

\usepackage{tikz}

\makeatletter
\patchcmd{\@makecaption}
  {\scshape}
  {}
  {}
  {}
\makeatother

\def\BibTeX{{\rm B\kern-.05em{\sc i\kern-.025em b}\kern-.08em
    T\kern-.1667em\lower.7ex\hbox{E}\kern-.125emX}}
\begin{document}

\title{Improving Tree-LSTM with Tree Attention \\
{\footnotesize \textsuperscript{}}
}

\author{\IEEEauthorblockN{Mahtab Ahmed, Muhammad Rifayat Samee, Robert E. Mercer}
\IEEEauthorblockA{\textit{Department of Computer Science},
\textit{University of Western Ontario},
London, Ontario, Canada \\
\{\texttt{mahme255, msamee, rmercer\}@uwo.ca}}}

\maketitle

\begin{abstract}
In Natural Language Processing (NLP), we often need to extract information from tree topology. Sentence structure can be represented via a dependency tree or a constituency tree structure. For this reason, a variant of LSTMs, named Tree-LSTM, was proposed to work on tree topology. In this paper, we design a generalized attention framework for both dependency and constituency trees by encoding variants of decomposable attention inside a Tree-LSTM cell. We evaluated our models on a semantic relatedness task and achieved notable results compared to Tree-LSTM based methods with no attention as well as other neural and non-neural methods and good results compared to Tree-LSTM based methods with attention.  
\end{abstract}


\section{Introduction}
Long Short Term Memory (LSTM) units are very effective when working on sequential data \cite{hochreiter1997long, gers1999learning}. For some Natural Language Processing (NLP) tasks, we often need to find a distributed representation of phrases and sentences \cite{le2014distributed, lin2017structured, dasgupta2018evaluating}. One obvious way of doing this is to use a sequential LSTM which captures word order in a sentence \cite{zhang2018sentence,palangi2016deep}. But we can also have information about sentence structure from a dependency parse tree or about phrase structure from a constituency tree \cite{klein2004corpus}. Despite the fact that RNN based models work well with sequence information, they frequently neglect to catch any sort of semantic compositionality if the information is structured rather than in the sequential frame \cite{socher2011parsing}. For example, the syntactic principles of natural language are known to be recursive, with noun phrases containing relative clauses that themselves contain noun phrases, e.g., \textit{``I went to the church which has nice windows''}\cite{socher2012semantic}. The term compositionality can also be explained in terms of a \textit{car}. A \textit{car} can be recursively decomposed into smaller \textit{car}  parts, for example, tires and windows and these parts can occur in different contexts, like tires on airplanes or windows in houses.

Attention \cite{bahdanau2014neural, luong2015effective, vaswani2017attention} was first introduced for doing machine translation  where the target word generated by the decoder at each time step is aligned with all the words in the source sentence. In its general form, attention allows a model to put importance on certain parts of the sentence for doing any specific downstream task \cite{yang2016hierarchical, du2018text}. In a dependency tree, the relationship between the entities (head and dependent) are organized as a structure where a head word can have multiple dependents under it. In the case of a constituency tree, a phrase is represented by one of the subtrees with the root being the phrase type and words or subtrees being the children. In both tree structured LSTMs, the derivation of the vector representation of the entire tree does not depend on all of the subtree components uniformly. Some parts of the tree have a larger influence on the root vector and some parts may have less. This contribution from subtrees for the building of the whole tree depends on the underlying task that the model is performing. For example, in a sentiment analysis task the sentiment of a tree depends on the sentiment of all of its children and how this information propagates. There may be scenarios where a single word (such as ``not'') flips the sentiment of the whole subphrase. These words should get more attention when deciding the sentiment of a subphrase containing them. On the other hand, when the problem is a regression problem with the task of assigning a score based on the semantic similarity of two sentences, this attention can be calculated as a cross sentence attention. In this case the representation of one sentence can guide the structural encoding of the other sentence on the dependency as well as constituency parse tree \cite{zhou2016modelling}.

Capturing semantic relatedness means recognizing the textual entailment between the hypothesis and the premise \cite{marelli2014sick}. The general approach of modeling sentence pairs (i.e., measuring the relatedness between sentences) using neural networks includes two steps: represent both of the sentences as vectors via a sentence encoder and then initializing a classifier with these vectors to do the classification. The sentence encoder can be viewed as a compositional function which maps a sequence of words in a sentence to a vector. Some of the common compositional functions are sequential LSTM \cite{zhou2016modelling}, Tree-LSTM \cite{tai2015improved, zhou2016modelling} and CNN \cite{he2015multi}.

In this paper, we propose two models to encode attention inside tree structured LSTM cells and verify their effectiveness by evaluating them on the semantic relatedness task where the model needs to give a score depending on how similar two sentences are. The tree data structure allows a set of dependents in the dependency tree or constituents in the constituency tree to be children of an immediately higher level (parent) tree node. Our tree attention model applies attention over the set of children in a subtree and decides which of them are important to reconstruct their parent node vector. We apply this attention with respect to four pieces of information: the vector representation of the sentence currently being represented as a tree, the vector representation of the sentence being compared with, dependent vectors (dependency tree) or phrase vector (binary constituency tree), and concatenated vectors of the dependents or the constituents. Our extensive evaluation proves the effectiveness of our attentive Tree-LSTM with respect to the plain Tree-LSTM models as well as some top performing models on the benchmark dataset.

\section{Related work}\label{relatedwork}
Socher et al.\ \cite{socher2011parsing} propose a number of recursive neural network (rNN) based models which take phrases as input rather than entire sentences. Phrases are represented as a vector as well as a parse tree. Vectors for higher level nodes in the tree are computed using a tensor-based composition function. Their best model was Matrix Vector rNN (MVrNN) where each word is represented as a vector as well as a matrix. In this model the children in a subtree interact more through their vectors rather than being influenced by some weights during the calculation of the parent's vector and matrix representation.

Tai et al.\ \cite{tai2015improved} developed two different variants of standard linear chain LSTMs: child sum Tree-LSTM and N-ary Tree-LSTM. The underlying concept of using input, output, update and forget gates in these variants is quite similar to how these gates are used in standard LSTMs, however there are few important changes. The standard LSTM works over the sequence data whereas these variants are compatible with tree structured data (constituency tree or dependency tree). Also, unlike standard LSTMs, the hidden and cell states of a word at the current time step does not depend on the entire sequence seen before. Instead, the hidden and cell state of a parent node depends only on its children hidden and cell states. Recently, Chen et al. \ \cite{chen2016enhanced} combined LSTM with Tree LSTM for natural language inference task and empirically proved that these two models complement each other very well.

Zhou et al.\ \cite{zhou2016modelling} extend the concept of standard Tree-RNNs and propose a number of attention based Tree-RNN models to perform the semantic relatedness task. Their insight was quite novel: in order to compute the semantic similarity of two sentences, one can encode attention in the tree structure of one sentence with respect to the vector representation of the other sentence. However, their proposed attention model only works with child sum Tree-LSTMs and GRUs. Attention with Tree LSTM has also been studied by Liu et al. \cite{liu2017attention} for text summarization task where they use two different kinds of alignment : block alignment for aligning phrases and word alignment for aligning inter-words within phrases.

Turning to machine translation, the attention mechanism is used to align the source and target sentences in the decoding phase. More formally, the attention mechanism allows the model to attend to some elements with the intention of emphasizing different elements. The well-known attention models from \cite{bahdanau2014neural} and \cite{luong2015effective} use recurrent models to attend over a set of source words during the generation of each target word. Using recurrent models to generate an attention score incorporates a memory mechanism inside the network which helps the model at run time to traverse and decide what to attend over. Also, this recurrency allows some positional information in the sequence to help ordering the generated words.

Parikh et al.\ \cite{parikh2016decomposable} propose a decomposable attention model for natural language inference tasks by removing the modules with recurrent behavior during the calculation of attention. First, they pick a single vector from a set of vectors representing the source sentence and then compare its point-wise similarity with every element of each word vector from the target sentence. Following this, they compare these alignments using a function which is a feed forward neural network and finally perform an aggregation through summation before doing the final classification. Gehring et al.\ \cite{gehring2017convolutional} propose a sequence to sequence learning framework utilizing a convolutional neural network which completely avoids recurrent models allowing their architecture to be parallelizable. In order to capture the positional information, they include a positional embedding layer which gives their model a sense of the portion of the sequence in the input or output it is currently dealing with. They encode \textit{sine} and \textit{cosine} frequencies for each dimension of every position in the sentence to create the positional embeddings and finally combine them with word embeddings. Vaswani et al.\ \cite{vaswani2017attention} combine the previous two works and propose a powerful machine translation framework utilizing attention without recurrence and positional embeddings. They also extend the decomposable attention mechanism by attending over the input sequence multiple times stating it as a multi-head attention where the target is to extract different features by different attentional heads.

\section{The Model}

In this section, we describe our work in detail. We first explain how the two variants of Tree-LSTM work. Following this, we describe our universal attention mechanism that is applicable for these two Tree-LSTM variants. Additionally, we give an in-depth analysis of adding this attention with respect to various information as discussed in Section \ref{relatedwork}.

\subsection{Incompatibility of standard LSTM and Tree structured data}

Recurrent neural networks (RNNs) are the best known and most widely used neural network (NN) model for sequence data as they sequentially scan the entire sequence and generate a compressed form of it. Although in theory RNNs are capable of remembering long distance dependencies, practically, as the sequence becomes longer, RNNs suffer from the 
vanishing gradient problem \cite{bengio1994learning, pascanu2013difficulty}. To overcome this drawback some RNN variants have been introduced such as Long Short Term Memory (LSTM) \cite{hochreiter1997long} and Gated Recurrent Unit (GRU) \cite{cho2014learning}. These variants use a gating mechanism to propagate new information further and at the same time to forget some previous information allowing the gradients to propagate further.
Performance-wise, LSTMs are superior to GRUs because they have more parameters but in terms of computational complexity GRUs often surpass LSTMs. 

Even though these gating variants effectively solve the RNN vanishing gradient problem, they are limited to linear data; however, a natural language sentence encodes more than a sequence of words. This extra information is usually represented in a tree structure. The tree structure 
shows how the words combine through different sub-phrases to reflect the overall meaning. If a sentence gets traversed by a standard LSTM, the latter part of the sentence gets more importance comparatively as the traversal moves left to right. But if the tree structure of the sentence gets traversed from bottom to top then the information from different constituent or dependents first gets combined to represent the root at the upper level and then this roots gradually gets traversed as children and combined to represent the root at next level and so on. So in both cases an LSTM cell will forget previous information which for plain LSTM, is related to the length of the sentence and for Tree-LSTM, is related to the depth of the tree. Also in plain LSTM, the hidden and cell state of a word at time step $t$ depends on hidden and cell state of all the words from time step $1 \ldots t-1$. But in Tree-LSTM, the hidden and cell state of a root word depends only on the hidden and cell state of all of its children rather than all the words before it. 

\begin{figure*}
\begin{subfigure}{.5\textwidth}
  \centering
  \includegraphics[width=.8\linewidth]{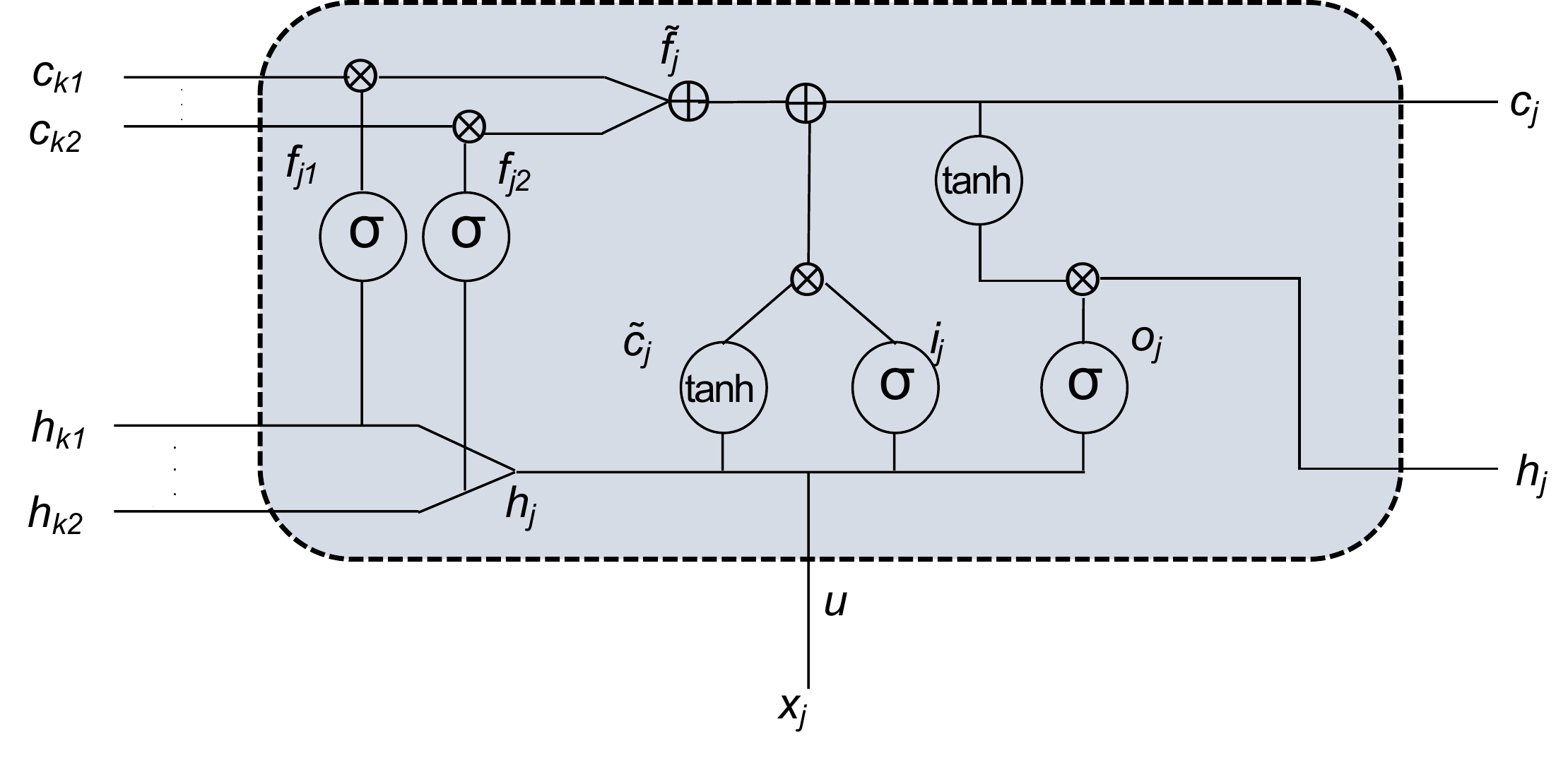}  
  \caption{Child Sum Tree-LSTM}
  \label{fig:sub-first}
\end{subfigure}
\begin{subfigure}{.5\textwidth}
  \centering
  \includegraphics[width=.8\linewidth]{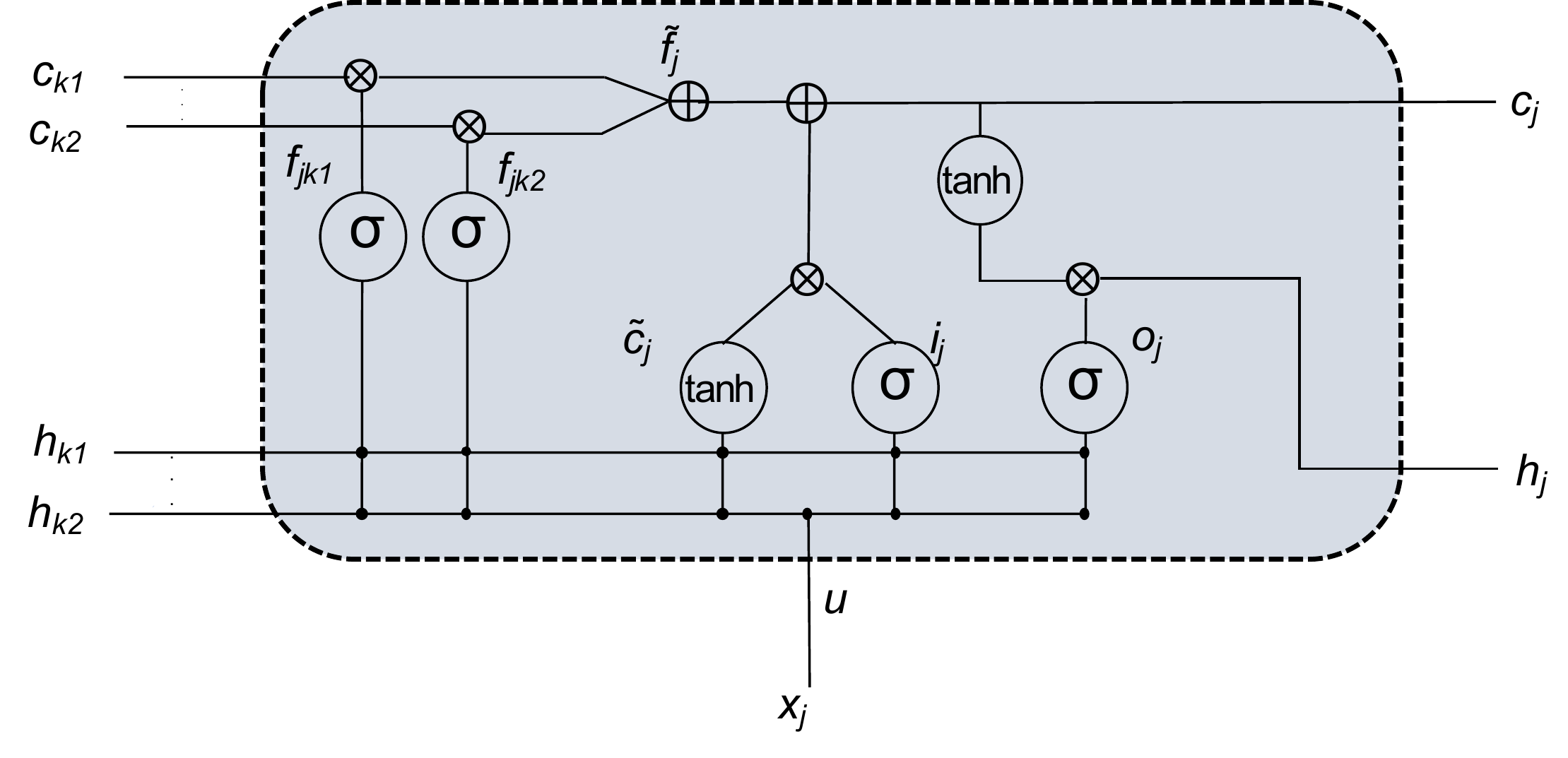}  
  \caption{Binary Tree-LSTM}
  \label{fig:sub-second}
\end{subfigure}

\begin{subfigure}{.5\textwidth}
  \centering
  \includegraphics[width=.8\linewidth]{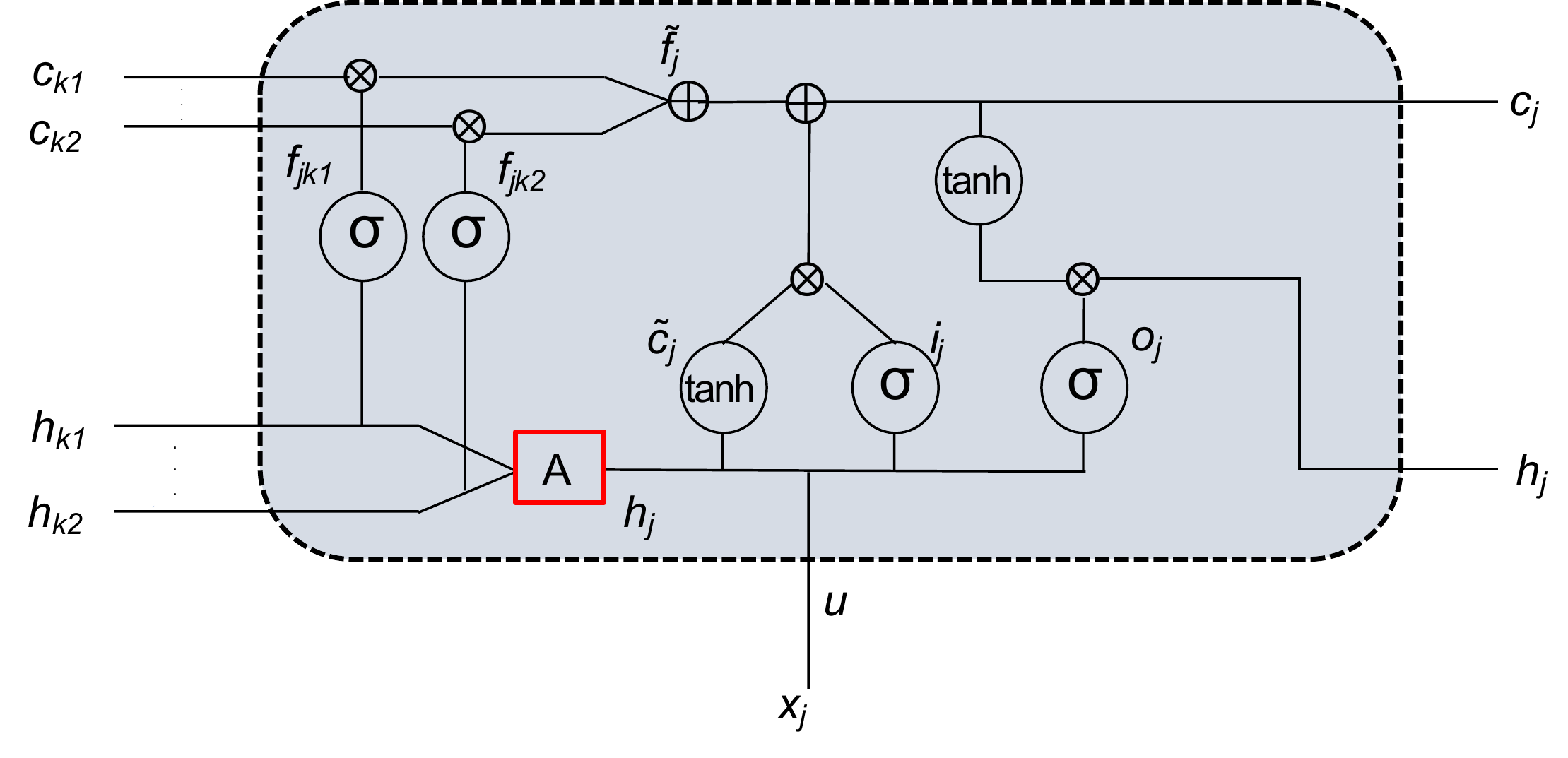}  
  \caption{Attentive Child Sum Tree-LSTM}
  \label{fig:sub-third}
\end{subfigure}
\begin{subfigure}{.5\textwidth}
  \centering
  \includegraphics[width=.8\linewidth]{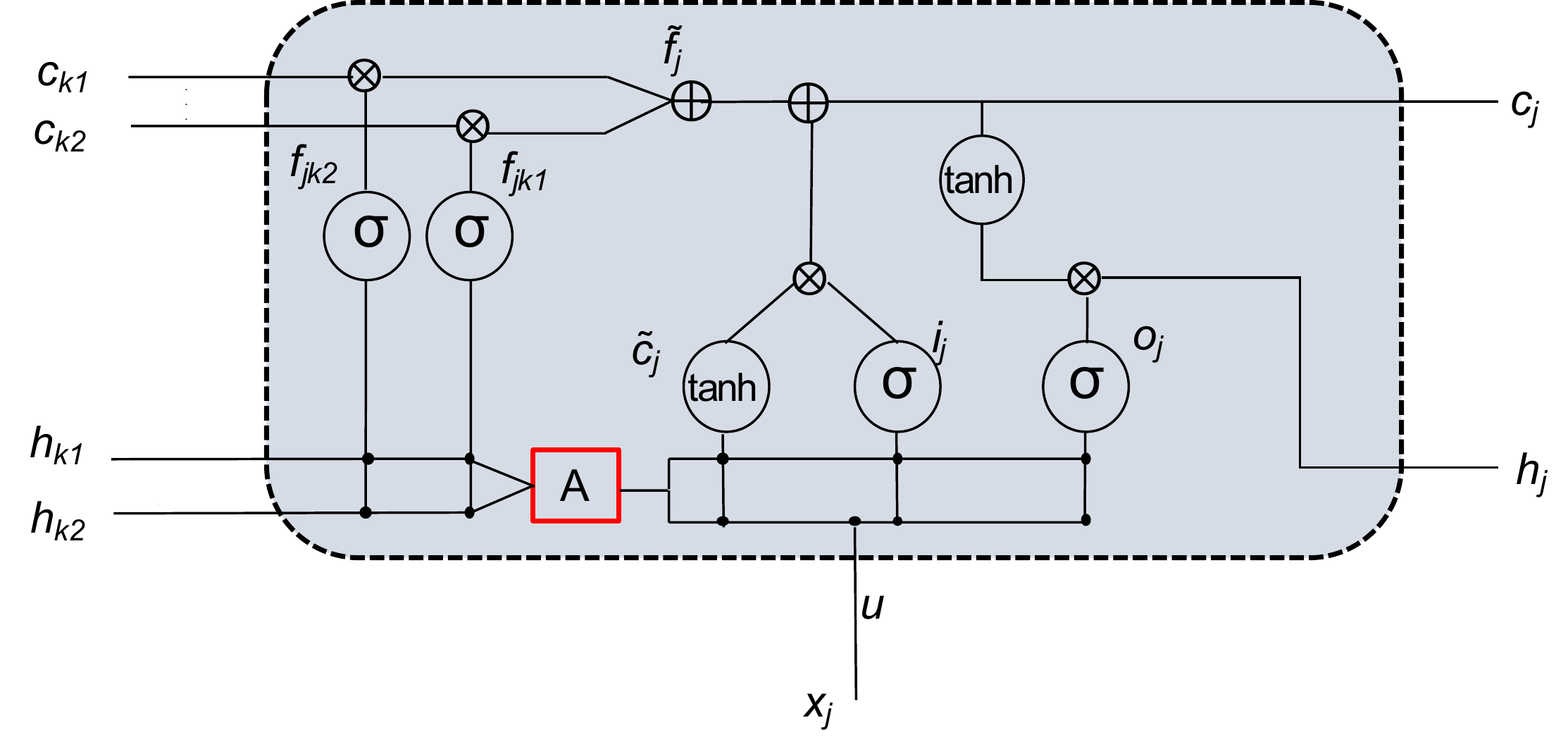}  
  \caption{Attentive Binary Tree-LSTM}
  \label{fig:sub-fourth}
\end{subfigure}
\caption{Illustrations of different Tree-LSTM architectures}
\label{fig:fig}
\end{figure*}
\subsection{Tree-LSTM}\label{treelstms}

There are two possible tree representations of a sentence: Dependency tree and Constituency tree \cite{chen2014fast}. As previously presented, the standard linear chain LSTM and BLSTM cannot correctly analyze this structured information. To properly deal with this structured data, Tai et al.\ \cite{tai2015improved} proposed two LSTM models which can analyze a tree structure preserving every property of the standard LSTM gating mechanisms. They called the first one child sum Tree-LSTM and the second one N-ary Tree-LSTM. Child sum Tree-LSTM fits well with dependency trees as it is well suited for high branching child-unordered trees. On the other hand N-ary Tree-LSTM (with $n=2$) works better with the binarized (Chomsky Normal Form) constituency trees. 

Traditional LSTM generates a new hidden and cell state from the previous hidden state $h_{t-1}$, previous cell state $c_{t-1}$ and current sequential input $x_t$. In the child sum Tree-LSTM, a component node state is generated based on the states of its children in the tree as shown in Fig.\ \ref{lstm}(a). To do this, the internal gates (i.e., the input, output and intermediate cell states) are updated using the sum of the hidden states of the children of the component node as follows: 
\begin{equation}\label{hsum}
    \Tilde{\textbf{h}}_{j} = \sum_{k \in C(j)}^{}h_{jk}
\end{equation}
where $C(j)$ denotes the children of node $j$. Next, using this modified hidden state, $\Tilde{h}$, the input, output and intermediate cell states are calculated as follows:
\begin{equation}
    \textbf{i}_{j} = \sigma (\textbf{W}^{(i)} x_{j} + \textbf{U}^{(i)} \Tilde{h}_{j} + \textbf{b}^{(i)})
\end{equation}
\begin{equation}
    \textbf{o}_{j} = \sigma (\textbf{W}^{(o)} x_{j} + \textbf{U}^{(o)} \Tilde{h}_{j} + \textbf{b}^{(o)})
\end{equation}
\begin{equation}
    \Tilde{\textbf{c}}_{j} = \textit{tanh} (\textbf{W}^{(c)} x_{j} + \textbf{U}^{(c)} \Tilde{h}_{j} + \textbf{b}^{(c)})
\end{equation}
where $W^{(i)}$, $W^{(o)}$, $W^{(c)}$, $U^{(i)}$, $U^{(o)}$, $U^{(c)}$, $b^{(i)}$, $b^{(o)}$, and $b^{(c)}$ are the parameters to be learned. Instead of having just a single forget gate, child sum Tree-LSTMs have $k$ forget gates where $k$ is the number of children of the target node. This multiple forget gate allows child sum Tree-LSTM to incorporate individual information from each of the children in a selective manner. Each forget gate is calculated as follows: 
\begin{equation}
    \textbf{f}_{jk} = \sigma (\textbf{W}^{(f)} x_{j} + \textbf{U}^{(f)} h_{jk} + \textbf{b}^{(f)})
\end{equation}
Next, the individual forget gate outputs are multiplied with corresponding cell state values and then combined to get a single forget vector which is used to get the final cell state of the model as follows: 
\begin{equation}
    \tilde{\textbf{f}}_{j} = \sum_{k \in C(j)}^{}f_{jk} \cdot {c_{k}}
\end{equation}
\begin{equation}\label{cell}
    \textbf{c}_{j} = i_{j} \cdot \tilde{c}_{j} + \tilde{f}_{j}
\end{equation}
Finally, the update equation for the hidden state of a child sum Tree-LSTM cell is similar to the traditional LSTM:
\begin{equation}\label{hidden}
    \textbf{h}_{j} = o_{j} \cdot \texttt{tanh}(c_{j})
\end{equation}

Each of the parameter matrices represents a correlation among the component vector, input $x_j$ and the hidden state $h_k$ of the $k^{th}$ child of the component unit. For example, the \texttt{sigmoid} function at the input gate represents semantically important words at input by giving values close to 1 (e.g., a verb) and relatively unimportant words by giving values close to 0 (e.g., a determiner).
Since the hidden state and cell state values of the parent node are generated based on the hidden state and the cell state of its children, child sum Tree-LSTM is well suited for dependency trees.

The N-ary Tree-LSTM is used where there are at most $N$ ordered children. Unlike the child sum Tree-LSTM, it has a different set of parameters for each child having its own cell and hidden state, shown in Fig.\ \ref{lstm}(b). The update equations for deriving input, output and update gate values are as follows:

\begin{equation}
    \textbf{i}_{j} = \sigma (\textbf{W}^{(i)} x_{j} + \sum_{l=1}^{N}\textbf{U}^{(i)}_l \Tilde{h}_{jl} + \textbf{b}^{(i)})
\end{equation}
\begin{equation}
    \textbf{o}_{j} = \sigma (\textbf{W}^{(o)} x_{j} + \sum_{l=1}^{N}\textbf{U}^{(o)}_l \Tilde{h}_{jl} + \textbf{b}^{(o)})
\end{equation}
\begin{equation}
    \Tilde{\textbf{c}}_{j} = \textit{tanh} (\textbf{W}^{(c)} x_{j} + \sum_{l=1}^{N}\textbf{U}^{(c)}_l \Tilde{h}_{jl} + \textbf{b}^{(c)})
\end{equation}
where $W^{(i)}$, $W^{(o)}$, $W^{(c)}$, $U^{(i)}_l$, $U^{(o)}_l$, $U^{(c)}_l$, $b^{(i)}$, $b^{(o)}$, and $b^{(c)}$ are the parameters to be learned. As can be seen, for each gate, the N-ary Tree-LSTM has a set of $N$ parameter matrices associated with the $N$ hidden states whereas the child sum Tree-LSTM has just one. Next, for each of the children, forget gate values are calculated separately, as done in the child sum Tree-LSTM as follows:

\begin{equation}
    \textbf{f}_{jk} = \sigma (\textbf{W}^{(f)} x_{j} + \sum_{l=1}^{N}\textbf{U}^{(f)}_{kl} h_{jl} + \textbf{b}^{(f)})
\end{equation}

Similar to the child sum Tree-LSTM, these new forget gate values are multiplied with corresponding cell state values and then summed to get the final values for the forget gate:

\begin{equation}
        \tilde{\textbf{f}}_{j} = \sum_{l=1}^{N}f_{jl} \cdot {c_{jl}}
\end{equation}

Finally, the cell state and new hidden state values are updated using Equations \ref{cell} and \ref{hidden}.

\subsection{Attention}

The two tree structured LSTM models described in Section \ref{treelstms} treat every word within a sub-tree with equal probability. More specifically, in an N-ary Tree-LSTM, every word contributes uniformly to the building of the higher-level constituent. Likewise, the child sum Tree-LSTM architecture suggests that, within a dependency tree branch, a head word influences all of its dependent words in a similar way. When viewing the tree as a semantic representation of a sentence, this may not be the case in many scenarios. For a constituency tree, if a sub-tree contains some negative sentiment words, then it is not always the case that the sentiment of that particular constituent is negative. If the negative sentiment word is preceded by a negation, then the higher-level constituent becomes semantically positive because of the location of the negation word. To capture this type of information, attention is applied over the sub-tree components to apportion the importance of each sub-tree component when building the entire tree either semantically or syntactically. In this study, we are interested in applying semantic attention over the sub-tree components to see how they contribute to building a sub-tree.  

Attentive Tree-LSTM was proposed by \cite{zhou2016modelling} for doing the semantic relatedness task. 
They state that the effect of semantic relevance could be implemented as part of the sentence representation construction process using a Tree-LSTM where each child should be assigned a different weight. In their proposed model, a soft attention mechanism assigns an attention weight on each child in a subtree. Given a collection of hidden state $h_1, h_2, \cdots, h_n$ and an external vector $s$, their proposed attention mechanism assigns a weight $\alpha_k$ on each of these hidden states and produces a weighted vector $g$. To achieve this, first they perform an affine transformation on each of the child hidden states and calculate a vector $m_k$ as follows:

\begin{equation}\label{mk}
    m_k = \texttt{tanh}(\textbf{W}^{(m)}h_k + \textbf{U}^{(m)}s),
\end{equation}
where $W^{(m)}$ and $U^{(m)}$ are the parameter matrices of size $d \times d$ and $s$ is the vector representation of the sentence learned by a sequential LSTM. Next, using this transformed hidden states $m_k$, the attention probabilities $\alpha_k$ are calculated as follows

\begin{equation}\label{alphak}
    \alpha_k = \frac{\texttt{w}^Tm_k}{\sum_{j=1}^{n}\texttt{w}^Tm_j}
\end{equation}
where $w$ is a parameter vector of size $1 \times d$. Following this, a weighted combination of the hidden states is calculated using,

\begin{equation}\label{g}
    g = \sum_{1 \leq k \leq n}^{}{\alpha_k}h_k
\end{equation}
This $g$ is of size $1 \times d$. Finally, an affine transformation is applied on this $g$ to get the new hidden state $\tilde{h}$ as follows:
\begin{equation}\label{htilde1}
    \tilde{h} = \texttt{tanh}(\textbf{W}^{(a)}g + \textbf{b}^{(a)})
\end{equation}

This soft attention mechanism from \cite{zhou2016modelling} introduces four new parameters to derive the final attentive hidden state; two matrices in Eqn. \ref{mk}, one vector in Eqn. \ref{alphak} and one matrix in Eqn. \ref{htilde1}. This attention mechanism is only applicable to the child sum Tree-LSTM. It is not possible to apply this attention on N-ary Tree-LSTMs since the structure of the N-ary Tree-LSTM is such that it needs $N$ separate hidden states to work with whereas a child sum Tree-LSTM collapses all the hidden states to a single vector through summation. In this study, we develop two generalized attention models by adopting the decomposable attention framework proposed by \cite{parikh2016decomposable} and the soft attention mechanism proposed by \cite{zhou2016modelling}.

\textbf{Model 1:} Our first model is based on the self attention mechanism where we make some subtle changes to calculate the attention probability with respect to different segments of the sentence. Calculating attention in this way involves three matrices \textit{key}, \textit{query}, and \textit{value}. The \textit{key} matrix represents on which child to attend over, the \textit{query} matrix represents ``with respect to what'' is attention to be applied and the \textit{value} matrix extracts the final attention-able vector using attention probability. The \textit{key} matrix is calculated as follows:

\begin{equation} \label{key}
    \textit{key} = \textbf{W}^{(k)}M^{(k)}
\end{equation}
where, $W^{(k)}$ is a parameter matrix of size $d \times d$ and $M^{(k)}$ is the matrix on which to attend over. For child sum Tree-LSTMs, this matrix is the concatenation of the vectors of all the words under a particular head word. For N-ary Tree-LSTMs, it is the concatenation of all the word vectors in a constituent. So in both cases the formal representation is $M^{(k)} = [h_1; h_2; \dots ; h_n]$. In order to encode self attention in the sub-tree, the \textit{query} and \textit{value} matrices also get calculated with respect to $M^{(k)}$ ($M^{(k)}=M^{(q)}=M^{(v)}$) but with a different set of parameter matrices $W^{(q)}$ and $W^{(v)}$ as follows:
\begin{equation}\label{query}
    \textit{query} = \textbf{W}^{(q)}M^{(q)}
\end{equation}
\begin{equation}\label{value}
    \textit{value} = \textbf{W}^{(v)}M^{(v)}
\end{equation}

Once the \textit{key} and \textit{query} get calculated, the next step is to align each of them by looking at the similarity at each dimension of their representation. This is done using:

\begin{equation}\label{align}
    \textit{align} = (\textit{query})^T\textit{key}\cdot\frac{1}{\sqrt{d}}
\end{equation}
where the \textit{align} matrix is of size $n \times n$ with $n$ representing the number of children within this sub-tree. The $d$ is being used here as a normalizing factor. Finally, the attention probability is calculated by applying \texttt{softmax} over it as follows:

\begin{equation}\label{alpha}
    \alpha = \texttt{softmax}(\textit{align}) 
\end{equation}

Here $\alpha$ is the matrix of attention probabilities where each row represents how much attention needs to be given on each of the children within that sub-tree according to the word at that row. As there are $n$ children within a sub-tree, the size of this matrix is $n \times n$. Finally, we calculate a new attention encoded hidden state $\tilde{h}$ through a batch-wise matrix multiplication between the $\alpha$ and \textit{value} matrices as follows:

\begin{equation}\label{htilde}
    \tilde{h} = \texttt{bmm}(\alpha, \textit{value})
\end{equation}

The shape of this new $\tilde{h}$ is $n \times d$. It contains attention encoded hidden state values of all the children sequentially one on top of another. So in order to locate a specific hidden state value, the row number corresponding to the position of that child in the sub-tree is used. For child sum Tree-LSTMs, all of the hidden state vectors are summed to get a single vector and for N-ary Tree-LSTMs, one row of $\tilde{h}$ is selected as the hidden state of a child.  

For the semantic relatedness task, where the objective is to assign a score based on the similarity between two sentences, it is better to calculate the query matrix with respect to the vector representation of the second sentence. Specially, given a pair of sentences, our generalized attentive encoder uses the representation of one sentence generated via a sequential LSTM to guide the structural encoding of the other sentence on both 
the dependency as well as the constituency tree. In that case, $M^{(q)}$ is a vector rather than a matrix thus changing the shape of \textit{query} from Eqn. \ref{query} into $1 \times d$. This results in an alignment vector from Eqn. \ref{align} of size $1 \times n$. When \texttt{softmax} is applied over this vector, a vector of probabilities, $\alpha$, is produced. Finally, instead of doing a matrix multiplication as in Eqn. \ref{htilde}, a point-wise multiplication $\tilde{h} = \alpha * \texttt{value}$ is performed resulting in a new hidden state vector. For child sum Tree-LSTMs, we use this new hidden state vector in place of the one generated in Eqn. \ref{hsum} and for N-ary Tree-LSTMs, we use this hidden state vector as the hidden state of both the left and right children.  This way of calculating self attention requires three additional matrices as parameters from Eqn. \ref{key}, \ref{query} and \ref{value}, a smaller number of parameters than found in \cite{zhou2016modelling}. We further continue our experiments by calculating a phrase vector representation using an additional LSTM cell and use it as the \textit{query} vector. Then, we adopt the same procedure as above to calculate the attention probability $\alpha$ and the final hidden state vector $\tilde{h}$. However, this requires more parameters than what is required in \cite{zhou2016modelling}.

\textbf{Model 2:} In our second model, we combine the concepts of decomposable attention mechanism with a soft attention layer. Here, we have two matrices \textit{key} and \textit{query} and their derivation are the same as Eqns. \ref{key} and \ref{query}. We further align and transform these matrices into probabilities using the same set of equations, Equations \ref{align} and \ref{alpha}. We again make some subtle changes which result in four different versions of this model. In Eqn. \ref{query}, when $M^{(q)} = M^{(k)}$, the dimension of the attention probability becomes $n \times n$ and when $M^{(q)}$ is either a sentence vector $M^{(q)} = \texttt{LSTM}(\textit{sentence}_2)$ or phrase vector $M^{(q)} = \texttt{LSTM}(M^{(k)})$, the dimension of this attention probability changes to $1 \times n$. Then, $\tilde{h}$ is calculated as follows,

\begin{equation}
    \tilde{h}  = 
    \begin{cases}
      \texttt{bmm}(\alpha, \textit{M}^{(k)}), & \text{if}\; \alpha is a matrix \\
      \alpha * \textit{M}^{(k)}, & \text{if}\; \alpha is a vector 
    \end{cases}
\end{equation}

Next we perform an affine transformation of this $\tilde{h}$ by multiplying it with a parameter matrix $W$ and passing it through a \texttt{tanh} layer as follows:

\begin{equation}
    \hat{h} = \texttt{tanh}(\textbf{W}\tilde{h} + \textbf{b})
\end{equation}

In the case of child sum Tree-LSTMs, if $\hat{h}$ is a matrix, we do a summation of all the rows and use that as the final vector and if $\hat{h}$ is a vector. we use that as it is. In the case of N-ary Tree-LSTMs, if $\hat{h}$ is a matrix, then each row corresponds to the hidden state of a child and if $\hat{h}$ is a vector, then we just copy this vector as the hidden states of the children.

\section{Experimental Setup and Analysis}
In  this  section,  we describe  the  detailed  experimental setup for the evaluation of our study. We first explain the dataset statistics for evaluating our generalized attention frameworks. Following this, we explain the working environment details along with the hyper-parameter settings of our architecture.

We evaluated our model for the semantic similarity task on the Sentences Involving Compositional Knowledge (SICK) dataset \cite{marelli2014sick}. The task is to give a likeness score for a pair of sentences and then compare it to a human produced score. The SICK dataset contains 9927 sentence pairs configured as: 4500 training pairs, 500 development pairs and 4927 test pairs. Each sentence pair is annotated with
a similarity score ranging from 1 to 5. A high score shows that the sentence pair is strongly related. All sentences are derived from existing image and video comment datasets. The assessment measures are Pearson's $\rho$ and mean squared error (MSE).

\begin{table}[h]
\caption{\label{hyper}  Ranges of different hyper-parameters searched during tuning. }
\centering
\small
\begin{tabular}{ p{1.5in} | p{1.2 in} } \hline 
\textbf{Hyper-parameter} & \textbf{Range Selected} \\ \hline 
Learning rate & 0.01 / 0.025 / 0.05 \\ \hline 
Batch size & 10 / 25 / 30 \\ \hline 
Momentum & 0.9 \\ \hline 
Memory dimension & 150 \\ \hline
MLP hidden dimension & 50 \\ \hline
Attention layer dimension & 150 \\ \hline
Dropout & 0.5 / 0.2 / 0.1 \\ \hline 
Word embedding size & 300  \\ \hline 
Gradient clipping & 5 / 20 / 50 \\ \hline 
Weight decay & $10^{-5}$\\ \hline
Learning rate decay & 0.05\\ \hline
\end{tabular}
\end{table}
Table \ref{hyper} shows the detailed hyper-parameter settings of our model. We trained our model on an Nvidia GeForce GTX 1080 GPU with `Adam', `SGD' and `Adagrad' optimizers. All of the results in the next section are reported using `Adagrad' as it was giving the best results. The `Learning rate decay' parameter was only used with the `SGD' optimizer. We used PyTorch 0.4 to implement our model under the Linux environment.
\begin{table}[t]
\centering
\caption{\label{cross}Test set results on the SICK dataset. The first group lists previous results, and the remainder are the results of  our models. We mark models
that we re-implemented with a $\dagger$.}
\begin{tabular}{|c|c|c|l|l|}
\hline
\multirow{10}{*}{\begin{tabular}[c]{@{}c@{}}Previous\\ Models\end{tabular}} & \multicolumn{2}{c|}{\textbf{Model}} & \textbf{$\mathbf{r}$} & \textbf{MSE} \\ \cline{2-5} 
 & \multicolumn{2}{l|}{ECNU \cite{zhao2014ecnu}} & 0.8414 & \multicolumn{1}{c|}{---} \\ \cline{2-5} 

 & \multicolumn{2}{l|}{Combine-skip+COCO \cite{kiros2015skip}} & 0.8655 & 0.2561 \\ \cline{2-5} 
 & \multicolumn{2}{l|}{ConvNet\cite{he2015multi}} & 0.8686 & 0.2606 \\ \cline{2-5} 
 & \multicolumn{2}{l|}{Seq-GRU \cite{zhou2016modelling}} & 0.8595 & 0.2689 \\ \cline{2-5} 
 & \multicolumn{2}{l|}{Seq-LSTM \cite{zhou2016modelling}} & 0.8528 & 0.2831 \\ \cline{2-5} 
 & \multicolumn{2}{l|}{Dep. Tree-GRU \cite{zhou2016modelling}} & 0.8672 & 0.2573 \\ \cline{2-5} 

 & \multicolumn{2}{l|}{Dep. Tree-GRU + Attn. \cite{zhou2016modelling}} & 0.8701 & 0.2524 \\ \cline{2-5} 
 
 & \multicolumn{2}{l|}{ \multirow{2}{*}{Const. Tree-LSTM \cite{tai2015improved}}}& 0.8582 & 0.2734 \\ \cline{4-5} & \multicolumn{2}{l|}{}& 0.8460 $\dagger$ & 0.2895 $\dagger$ \\ \cline{2-5}
 
 & \multicolumn{2}{l|}{ \multirow{2}{*}{Dep. Tree-LSTM \cite{tai2015improved}}}& 0.8676 & 0.2532 \\ \cline{4-5} & \multicolumn{2}{l|}{}& 0.8663 $\dagger$ & 0.2612  $\dagger$ \\ \cline{2-5} 
 
 & \multicolumn{2}{l|}{ \multirow{2}{*}{Dep. Tree-LSTM + Attn. \cite{zhou2016modelling}}}& 0.8730 & 0.2426 \\ \cline{4-5} & \multicolumn{2}{l|}{}& 0.8635 $\dagger$ & 0.2591 $\dagger$ \\ \cline{2-5}
 
\hline  \hline

\multicolumn{1}{|l|}{\multirow{8}{*}{\begin{tabular}[c]{@{}c@{}}Child \\Sum\\ Tree\\
LSTM\end{tabular}}} & \multirow{4}{*}{Model 1} & Self & 0.7466 & 0.4545 \\ \cline{3-5} 
\multicolumn{1}{|l|}{} &  & Sentence 1 & 0.7305 & 0.4849 \\ \cline{3-5} 
\multicolumn{1}{|l|}{} &  & Sentence 2 & 0.7939 & 0.3801 \\ \cline{3-5} 
\multicolumn{1}{|l|}{} &  & Phrase & 0.7889 & 0.3877 \\ \cline{2-5} 
\multicolumn{1}{|l|}{} & \multirow{4}{*}{Model 2} & Self & \multicolumn{1}{l|}{0.8577} & \multicolumn{1}{l|}{0.2695} \\ \cline{3-5} 
\multicolumn{1}{|l|}{} &  & Sentence 1 & 0.8620 & 0.2634 \\ \cline{3-5} 
\multicolumn{1}{|l|}{} &  & Sentence 2 & 0.8686 & 0.2518 \\ \cline{3-5} 
\multicolumn{1}{|l|}{} &  & Phrase & 0.8623 & 0.2615 \\ \hline
\multirow{8}{*}{\begin{tabular}[c]{@{}c@{}}Binary\\ Tree\\  LSTM\end{tabular}} & \multirow{4}{*}{Model 1} & Self & 0.8648 & 0.2567 \\ \cline{3-5} 
 &  & Sentence 1 & 0.8692 & 0.2486 \\ \cline{3-5} 
 &  & Sentence 2 & 0.8686 & 0.2507 \\ \cline{3-5} 
 &  & Phrase & 0.8676 & 0.2517 \\ \cline{2-5} 
 & \multirow{4}{*}{Model 2} & Self & 0.8698 & 0.2476 \\ \cline{3-5} 
 &  & Sentence 1 & 0.8698 & 0.2476 \\ \cline{3-5} 
 &  & Sentence 2 & 0.8720 & 0.2435 \\ \cline{3-5} 
 &  & Phrase & 0.8696 & 0.2479 \\ \hline
\end{tabular}
\end{table}

Table \ref{cross} shows the overall evaluation of our model in terms of Pearson's $\rho$ and Mean Squared Error (MSE). This table also contains the results of some top performing models on the SICK dataset. Among these models, \cite{tai2015improved} and \cite{zhou2016modelling} did their evaluation with plain Tree-LSTMs, whereas the rest of the models use some different composition functions such as CNN \cite{kiros2015skip}, ECNU \cite{zhao2014ecnu} and Combine-skip + COCO \cite{he2015multi}. However \cite{zhou2016modelling} also experimented with attentive Tree-LSTMs and GRUs, but they have only been able to design models compatible with the child sum variant. On the other hand, among our two proposed models, Model 2 performs very well on both Tree-LSTM variants showing significant improvements with every configuration. For both child sum as well as binary Tree-LSTMs, our second model with cross sentence attention has superior performance compared to the plain Tree-LSTM variants getting MSE of $0.2518$ and $0.2435$ respectively. For the child sum Tree-LSTM, Model 1 performs poorly compared to all the other models. This poor performance  is due to the hard attention that it applies. If a subtree has $n$ children, this hard attention forces $n-1$ children to have probability close to $0$ which causes the domination of just one child hidden state in the summation. The rest will not contribute at all. On the other hand, the reason behind Model 2 performing better in every configuration with both variants is that even though a hard attention causes one of the children to get close to $0$, the normalization of N-ary tree into binary tree causes much more flexibility for the information to flow from bottom to top. During normalization, a branch with $n$ children gets split up to $n-1$ full binary trees resulting in $(n-1)/2$ nodes that are always chosen. Our best performing attentive child sum Tree-LSTM model with cross sentence attention achieves a better result ($0.2518$ MSE) than the plain child sum tree variant from \cite{tai2015improved} ($0.2532$ MSE). Our score did not surpass the reported result ($0.2426$ MSE) of the attentive child sum variant from \cite{zhou2016modelling}. However, our implementation of their model with their reported hyper-parameters gave a $0.2591$ MSE which is significantly worse than their claimed MSE. This suggests to us that the implementation environment has a strong impact on model performance. Our child sum Tree-LSTM Model 2 with cross sentence attention achieves better performance than our implementation of \cite{zhou2016modelling} using their hyper-parameter settings. To the best of our knowledge, our work is the first to encode attention inside a binary Tree-LSTM cell. In terms of binary tree LSTM, our best performing model with cross sentence attention achieves $0.2435$ MSE which is significantly better than the one reported in \cite{tai2015improved} ($0.2734$ MSE) for the non-attentive version. In our implementation of plain binary Tree-LSTM without attention from \cite{tai2015improved} we were not able to reproduce their reported result and ended up with $0.2895$ MSE which is much worse than the one we got with every configuration of our Model 1 and Model 2. This performance analysis does show the effectiveness of our generalized attention model.

\begin{figure}[b]
\centerline{\includegraphics[scale = 0.8]{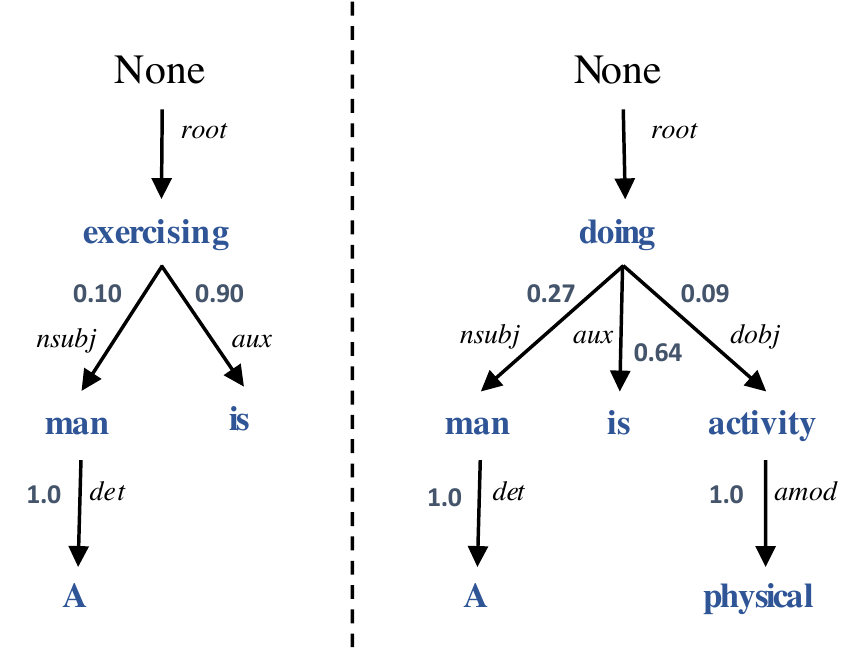}}
\caption{ \label{depend}Probability of each node being selected by attentive child sum Tree-LSTM Model 2 with cross sentence attention (\textit{\textbf{Left}: A man is exercising
\textbf{Right}: A man is doing physical activity
\textbf{Label}: Entailment}
).}
\end{figure}

\begin{figure}[ht]
\centerline{\includegraphics[scale = 0.7]{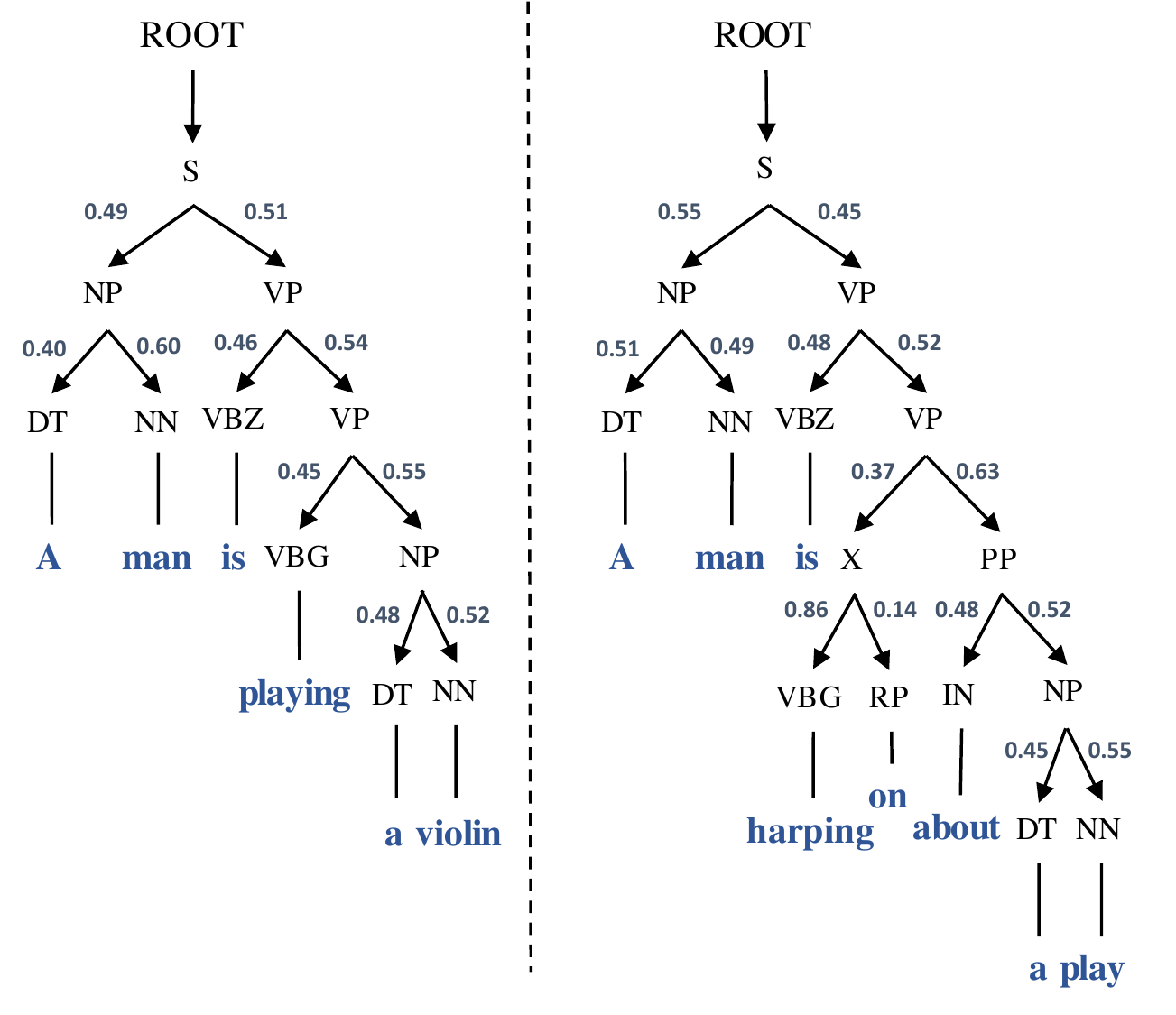}}
\caption{ \label{const}Probability of each node being selected by attentive binary Tree-LSTM Model 2 with cross sentence attention (\textit{\textbf{Left}: A man is playing a violin
\textbf{Right}: A man is harping on about a play
\textbf{Label}: NEUTRAL}
).}

\end{figure}

Figure \ref{depend} depicts the probability assigned to each node in the dependency tree by our Model 2 with cross sentence attention. Unlike standard child sum Tree-LSTM, where the hidden states of all the children nodes are combined with a plain summation, our attentive child sum Tree-LSTM assigns a weight to each node and then does a weighted summation. The example used in this figure has ``\textit{A man is exercising}" as the left sentence, ``\textit{A man is doing physical activity}'' as the right sentence and ``\textit{Entailment}" as their relationship. As usual, the main verb from both of the sentences is selected as the \textit{root} node. The auxiliary verb (\textit{is}) gets high attention in both the left and right trees because of the word similarity. However, their absolute influence varies because of the presence of semantically related words in other branches as discussed above. Both of these trees share the same nominal subject (\textit{nsubj}) however with different probabilities (in the left tree its probability is significantly lower). The reason behind this is the cross sentence attention allows the word \textit{man} from the left sentence to align with two words \textit{man} and \textit{physical} from the right sentence. As they share a similar semantic meaning in the vector space, the branch in the left sentence that contains \textit{man} is diminished because the right sentence divides the attention between two branches (left sentence: \textit{exercising} $\xrightarrow[]{\textit{nsubj}}$ \textit{man}; right sentence: \textit{doing} $\xrightarrow[]{\textit{nsubj}}$ \textit{man} and \textit{doing} $\xrightarrow[]{\textit{dobj}}$ \textit{activity}). 

Figure \ref{const} depicts the probability assigned to each node in a binary (Chomsky Normal Form) constituency tree using an attentive binary Tree-LSTM with cross sentence attention. In this setting, the attention on the structure of the left sentence is computed with respect to the vector representation of the right sentence and vice versa. As a result, the words in a specific phrase from the left sentence are aligned with very high attention probability if the same words appear anywhere in the right sentence. However, as \texttt{softmax} was operating with small values from Eqn. \ref{align}, it forced both children to have the same probabilities ($0.5$). In order to verify whether this probability has any effect or not, we have confirmed that replacing $\alpha$ in Eqn.\ \ref{htilde} with pairs of the same value other than $(0.5$) results in the model giving comparatively poor performance. Finally, for the inference of attention probabilities, we replaced \texttt{softmax} from Eqn.\ \ref{alpha} with plain normalization. For the example in Fig.\ \ref{const}, we have ``\textit{A man is playing a violin}'' as the left sentence, ``\textit{A man is harping on about a play}'' as the right sentence and ``\textit{Neutral}'' as their relationship. The phrase \textit{NP} gets almost the same probabilities in both the left ($0.49$) and right ($0.55$) trees because of having the same set of words: ``\textit{A man}''. The sub-phrase \textit{VBZ} under \textit{VP} in both trees gets very high attention due to having the same word ``\textit{is}'' at exactly the same position. Due to the Chomsky normalization, the tree on the right side gets an extra dummy node \textit{X} which contains \textit{VBG} and \textit{RP} as the child nodes. In the vector space, the words ``playing'' and ``harping'' are semantically connected which allows both of the models to align them with moderately high as well as equal probabilities. The left tree does not have any particle (``\textit{RP}'') words which causes the model to put low attention probability when it appears on the right tree. The left tree has \textit{NP} as the right child of \textit{VP} at level 3 with probability $0.55$ which is quite close to the amount of attention \text{PP} gets ($0.63$) as the right child of \textit{VP} at the same level in the right tree. Again in both of these trees, at the right most branch, the words ``play'' and ``violin'' share the same semantic space which causes them to get aligned with almost the same probabilities. The \textit{DT} in this branch gets the same high probability because of appearing in both sentences at relatively similar positions.

\section{Conclusion}
Previous attempts to encode the attention mechanism in Tree-LSTMs were only successful for the child-sum tree variant as the techniques used are not easily adaptable to binary trees like the Chomsky Normal Form constituency tree. In this paper, we have introduced two different ways of applying attention on tree structures. The second of these two methods gives superior performance for both tree variants. The proposed techniques can be used on both dependency as well as constituent tree structure. Our experimental results verify the superiority of the attentive variant of Tree-LSTMs over traditional Tree-LSTMs and linear chain LSTMs on the semantic relatedness task. With our extensive in depth analysis, we showed that our proposed attention models provide a good representation of how a sentence builds semantically from the words. Our generalized attention framework is adaptable to any tree like structures.

\bibliographystyle{IEEEtran}
\bibliography{reference}
\end{document}